\pdfoutput=1

\documentclass[10pt,twocolumn,letterpaper]{article}

\usepackage[pagenumbers]{cvpr}      %

\usepackage{float}      %
\usepackage{multirow}   %
\usepackage{adjustbox}  %
\usepackage{tikz}       %

\definecolor{cvprblue}{rgb}{0.21,0.49,0.74}
\usepackage[pagebackref,breaklinks,colorlinks,allcolors=cvprblue]{hyperref}

\def\confName{CVPR}
\def\confYear{2027}

\title{EventSSM: Causal Spatial-State Modelling for Asynchronous Events Event-Depth Estimation}

\author{Sadiq O. Macaulay\\
University of Surrey\\
{\tt\small s.macaulay@surrey.ac.uk}
\and
Dipon Kumar Ghosh\\
University of Surrey\\
{\tt\small d.ghosh@surrey.ac.uk} 
\and
Hadfield Simon\\
University of Surrey\\
{\tt\small s.hadfield@surrey.ac.uk}\\
}

\makeatletter
\apptocmd{\@maketitle}{\begin{center}
\begin{minipage}{\textwidth}
\centering
\iftoggle{cvprfinal}{\vspace*{-14pt}}{}%
\includegraphics[width=0.7\linewidth]{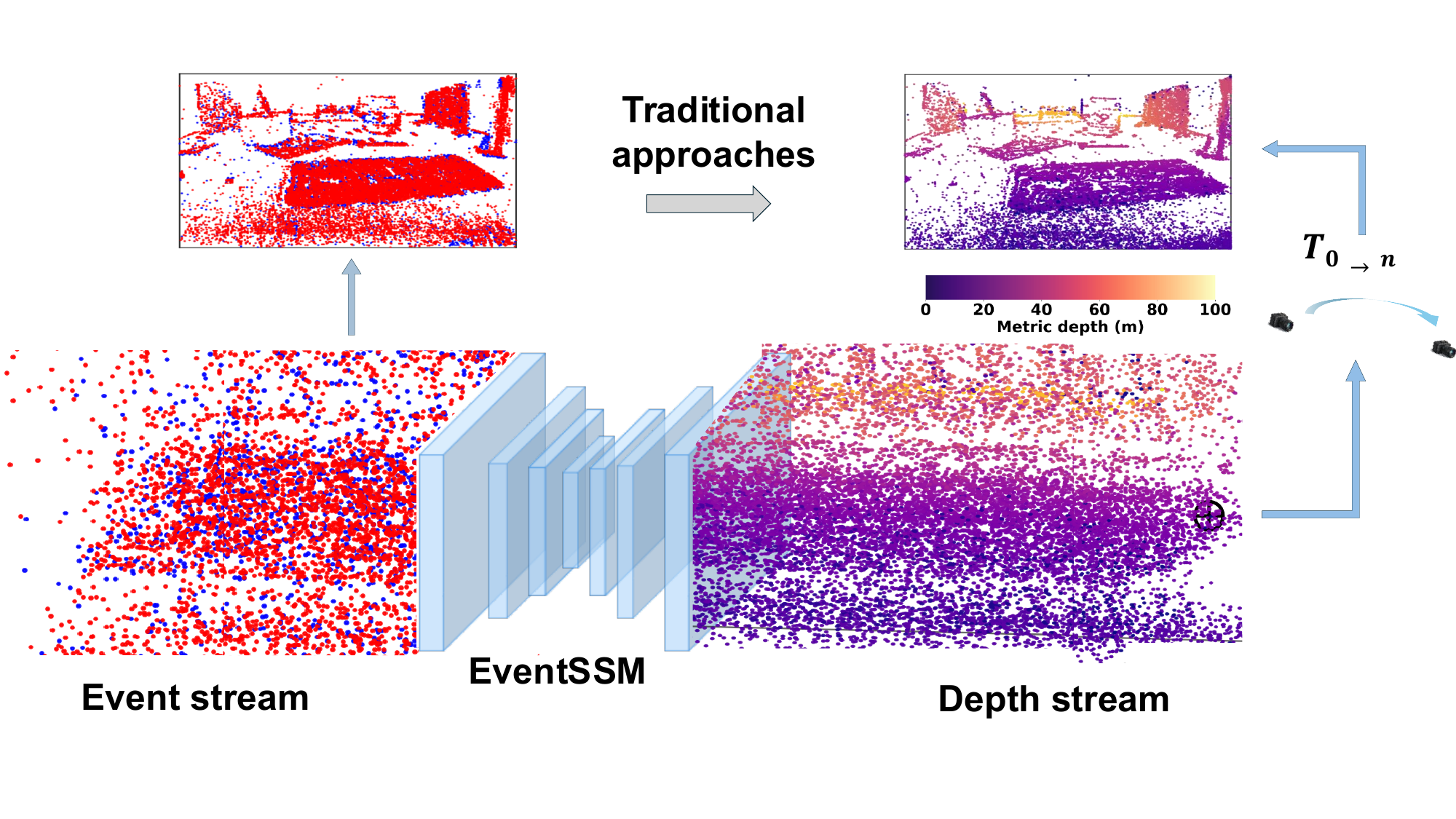}
\vspace{-1cm}
\captionof{figure}[EventSSM overview]{Overview of event-associated depth estimation and target-time geometric rendering from asynchronous observations.}
\label{fig:ch3:overview}
\end{minipage}
\end{center}
}{}{}
\makeatother

\begin{document}
\maketitle
\begin{abstract}

Event cameras report asynchronous brightness changes, yet metric-depth supervision is typically available only at sparsely sampled reference timestamps. Aggregating events into frame-like representations simplifies learning but can obscure the fine-grained temporal structure of the original event stream. We introduce \emph{EventSSM}, a causal spatial-state framework for event-associated metric-depth estimation using a recurrent memory updated by chronological event packets. To bridge asynchronous predictions and frame-time supervision, predicted event depths are geometrically transported using calibrated camera poses and differentiably rendered at reference timestamps. Benchmark evaluations demonstrate improvements over established event-depth methods across multiple depth-error measures. Controlled ablations further examine cross-modal initialization, temporal recurrence, individual state, readout and supervision components, and packet subdivision. Together, these findings establish EventSSM as an effective framework for causal event-associated metric-depth estimation, combining persistent spatial context with physically timed state retention.

\end{abstract}
    
\section{Introduction}
\label{sec:intro}
Event cameras report changes in logarithmic brightness as asynchronous, polarity-labelled events rather than periodically acquired intensity frames~\cite{Lichtsteiner2008DVS,Posch2014RetinomorphicEvent,Gallego2022EventSurvey}. Their fine temporal resolution and high dynamic range have supported high-speed optical flow, camera tracking, and geometric reconstruction~\cite{Bardow2016SimultaneousOptical,Rebecq2018EVO,Mueggler2018ContinuousTime}. However, an event specifies neither scene depth nor an image-wide geometric measurement: event activity depends on motion, local contrast, and sensor characteristics~\cite{Mueggler2017eventcamera,Gallego2018Contrast}. Repeated observations may concentrate along moving texture boundaries while nearby surfaces produce few events. Estimating depth therefore requires spatial context that persists across irregularly timed measurements.

A widely adopted strategy converts event streams into count images, time surfaces, or voxel grids before applying a dense predictor~\cite{Gehrig2019EndToEnd,Gehrig2023RecurrentVision}. This has enabled recurrent monocular depth networks such as E2Depth and recurrent-transformer estimators, as well as event--frame fusion~\cite{Hidalgo2020E2Depth,Liu2024RecurrentTransformers,Gehrig2021RAMNet}. More recent depth estimators exploit adverse-condition modelling or spatial knowledge transferred from image networks~\cite{Shi2023EVENEvent,Bartolomei2025DepthAnyEvent,Zhu2025DepthAnyEventStream}. These approaches provide important dense-depth baselines, but the acquisition window or temporal binning of an intermediate representation constrains when its prediction can be formed. For a model that attaches depth to the original event locations, a complementary design is to retain a spatial state between packets and query it using each event's coordinates, polarity, and age.

A second challenge concerns supervision. Event timestamps are much finer than the depth and pose annotations available in datasets such as MVSEC~\cite{Zhu2018MVSEC}; geometric reconstruction and continuous-time event-camera odometry both demonstrate the importance of accounting for camera motion between observations~\cite{Rebecq2018EMVS,Mueggler2018ContinuousTime}. Assigning a reference-frame depth directly to an event at another time ignores the change in viewpoint, although over short MVSEC intervals this has little effect on accuracy (Sec.~\ref{sec:exp-components}). We instead associate predictions with packet-time events, transport them through calibrated, interpolated camera poses, and render depth at the reference timestamp. This uses frame-time labels for learning and evaluation without forcing the prediction domain to become frame-synchronous.

We introduce \textbf{EventSSM}, an event-based encoder architecture that pairs a \emph{motion-transported spatial state} with a \emph{local event-conditioned depth readout}. Chronological packets produce polarity-separated observations that update a single spatial memory through learned backward transport, physical-time decay, and support-gated injection. A feed-forward multiscale pyramid supplies each event query with 20 spatial descriptors sampled from a fixed local neighborhood across four feature scales. Attending to these descriptors enables localized depth inference from the accumulated spatiotemporal context without pairwise attention across the event stream. The recurrence is causal at \emph{packet boundaries}: it can use later events \emph{within} the current packet, but not observations from subsequent packets. State-space sequence modelling motivates the persistent-state perspective~\cite{Gu2021S4,Gu2023Mamba}, while image-trained representations offer spatial initialization during training only~\cite{Ch5Simeoni2025DINOv3}.

Our contributions are: \textbf{(i)} a physically timed spatial-state transition that couples learned feature transport with observation-dependent updates; \textbf{(ii)} a multiscale, constant-neighborhood attention readout producing bounded metric depth at individual event coordinates; and \textbf{(iii)} a calibrated target-time renderer that places event-time depth predictions in the annotated reference view for supervision and matched evaluation. We evaluate these components on matched MVSEC outdoor windows through depth comparisons and controlled initialization, recurrence, component and packet-subdivision studies.
\section{Related Work}
\label{sec:related}
\paragraph{Event representations and temporal context.}
Count images, time surfaces, and event voxels expose asynchronous measurements to conventional convolutional networks; learned event encodings can optimise this conversion for downstream tasks~\cite{Gehrig2019EndToEnd,Gallego2022EventSurvey}. Image reconstruction provides another interface to frame-based vision~\cite{Rebecq2019EventsToVideo}, whereas point- and graph-based processing preserves irregular event support more explicitly~\cite{Schaefer2022AEGNN}. Recurrent event transformers extend temporal context across discretised inputs~\cite{Gehrig2023RecurrentVision,Liu2024RecurrentTransformers}. These methods motivate different trade-offs between spatial completeness, event timing, and the cost of irregular computation. EventSSM retains packet chronology in a persistent image-plane state but assigns its final depth estimates to events rather than to every image pixel.

\paragraph{Depth estimation from events.}
Classical event-based geometry exploits calibrated stereo or motion-induced multi-view consistency~\cite{Xie2017EventStereo,Rebecq2018EMVS}, while event-only monocular depth learning commonly produces dense maps at designated times~\cite{Hidalgo2020E2Depth,Liu2024RecurrentTransformers}. RAMNet combines event and frame observations in a recurrent depth architecture~\cite{Gehrig2021RAMNet}; other methods target adverse night-time conditions, occlusion handling, or dense-to-sparse transfer~\cite{Shi2023EVENEvent,Zhou2024Eventdepth,Zhu2025DepthAnyEventStream}. Depth AnyEvent investigates transferring visual knowledge to event-depth models~\cite{Bartolomei2025DepthAnyEvent}, and ScaleEvent extends event-stream pretraining with visual foundation-model representations~\cite{Ch5Chen2026ScaleEvent}. We compare against the corresponding event-depth baselines under matched reference-time evaluation. Our distinction lies primarily in the \emph{output domain}: the model predicts event-associated metric depth and renders those samples into a reference image for comparison, rather than treating a dense frame as its native depth prediction.

\paragraph{Continuous-time geometry and supervision.}
Motion compensation and contrast maximisation exploit the timestamp structure of events in flow and motion estimation~\cite{Gallego2018Contrast,Shiba2024SecretsEvent}. Event-based multi-view reconstruction and continuous-time visual-inertial odometry further demonstrate that camera geometry must be evaluated at the appropriate observation time~\cite{Rebecq2018EMVS,Mueggler2018ContinuousTime}. Event-based self-supervised approaches obtain learning signals through motion or cross-modal consistency~\cite{Zhu2019Unsupervised,Zhu2024EMoDepth,Hagenaars2025OnDevice}. Our training setting is different: metric reference depth and annotated camera poses are available. We use those poses to project event-associated 3D samples into the labelled viewpoint, addressing the temporal mismatch between irregular predictions and sparse metric supervision.

\paragraph{State-space models and spatial context.}
Structured and selective state-space models show how long-range sequential context can be maintained without all-pairs token attention~\cite{Gu2021S4,Gu2023Mamba}. Visual adaptations, including Vision Mamba, VMamba and LocalMamba, investigate spatial scans and locality for image features~\cite{Zhu2024VisionMamba,Liu2024VMambaVisual,Huang2025LocalMambaVisual}. These are related motivations rather than interchangeable implementations: EventSSM does not use a global Mamba scan. It maintains a full-resolution spatial tensor whose transition explicitly depends on elapsed physical time, learned image-plane transport and local event support. Its four-scale pyramid is constructed for readout and does not form an independently recurrent hierarchy.

\paragraph{Cross-modal spatial priors.}
Image-based depth models demonstrate the value of broad visual representations for monocular geometry~\cite{Ranftl2020TowardsRobust,Piccinelli2024UniDepth,Yang2024DepthAnythingV2}. Distillation into event-depth estimators offers one route to exploiting such priors without requiring RGB frames during deployment~\cite{Bartolomei2025DepthAnyEvent,Zhu2025DepthAnyEventStream}. Our initialization instead aligns intermediate event-state features with a frozen DINOv3 image encoder~\cite{Ch5Simeoni2025DINOv3}. The teacher participates only during representation initialization; the trained depth head and inference pipeline consume events and recurrent memory.
\section{Method}
\label{sec:method}
\begin{figure*}[!t]
\centering
\includegraphics[width=0.95\linewidth]{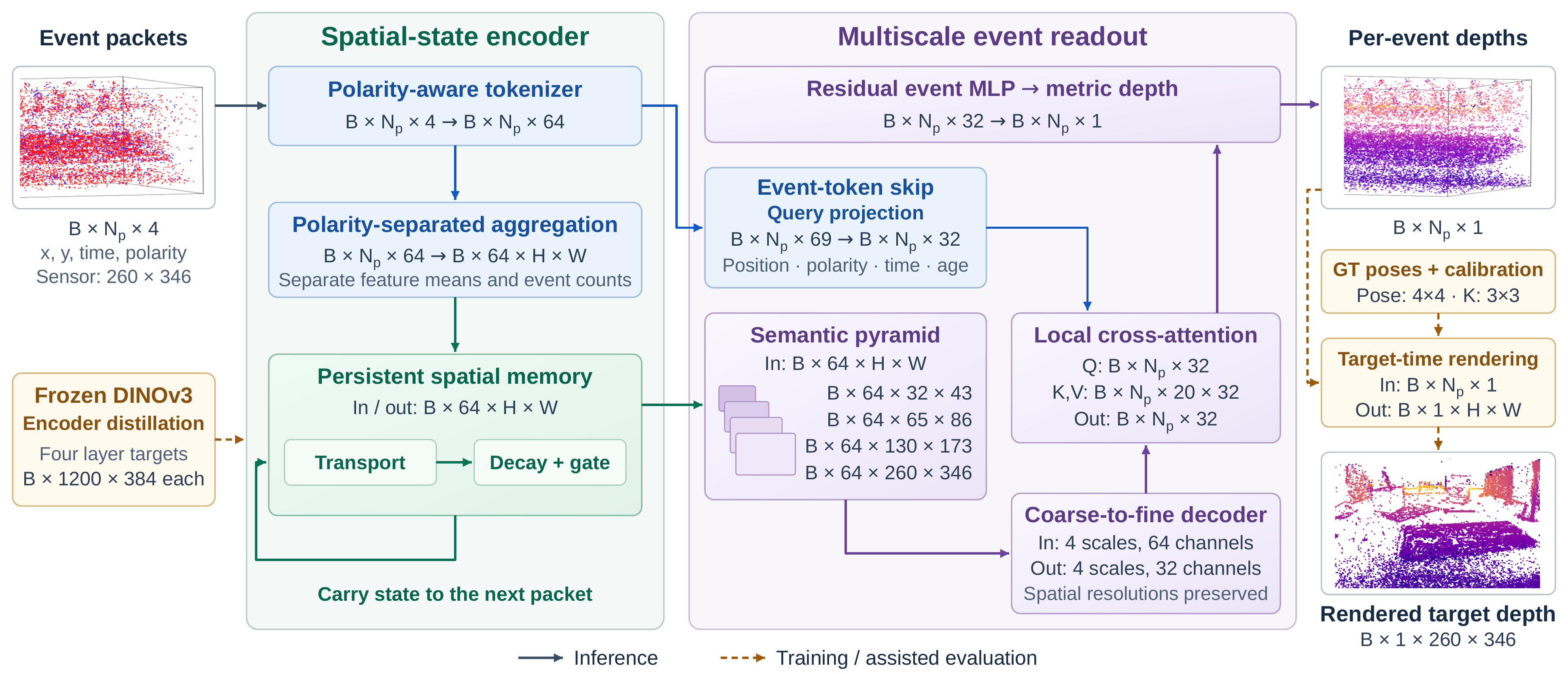}
\caption[Event-native spatial-state depth architecture]{Causal event-depth architecture. Each event packet updates a transported spatial memory; a derived multiscale feature pyramid provides descriptors for local event-conditioned cross-attention. The event-depth stream can be geometrically rendered at annotated reference times for supervision.}
\label{fig:ch3:continuous-depth-architecture}
\end{figure*}

Figure~\ref{fig:ch3:continuous-depth-architecture} illustrates the packet update and event-conditioned readout.

\subsection{Packet-causal event-depth prediction}
\label{sec:method-formulation}
An event stream is $\mathcal E=\{e_i=(x_i,y_i,t_i,p_i)\}_{i=1}^{N}$, where $p_i\in\{-1,+1\}$. We partition each accepted reference interval into $K$ ordered, non-overlapping packets $\mathcal E_k$ of duration $\Delta t_k$. Packet event counts vary with activity. The recurrent model returns one optical-axis depth per valid event:
\begin{equation}
M_k=F_\theta(M_{k-1},\mathcal E_k,\Delta t_k),\quad
\hat{\boldsymbol d}_k=G_\theta(M_k,\mathcal E_k).
\label{eq:state-interface}
\end{equation}
The output for $\mathcal E_k$ is available only after that packet closes. Its events can jointly affect $M_k$, while later packets cannot. This retains event-specific output coordinates, in contrast to conventional temporally accumulated event representations~\cite{Gehrig2019EndToEnd,Liu2024RecurrentTransformers}. A validity mask removes padded events from both the spatial update and readout.

\subsection{Polarity-aware spatial observations}
\label{sec:tokenizer}
Normalized coordinates, within-packet time, and polarity are embedded into a 64-dimensional event token $q_i$. For each pixel $u$ and polarity $p$, the spatial stem collects token means $\mu_k^p(u)$ and log-counts $\log(1+n_k^p(u))$. Concatenating the two mean maps and two count channels gives 130 features, projected to a 64-channel packet observation $O_k$ by pointwise mixing and a depthwise spatial convolution. The support indicator $S_k(u)=\mathbb{1}[n_k^+(u)+n_k^-(u)>0]$ identifies pixels at which a new observation may enter the recurrent memory. The polarity-specific pooling preserves signed contrast information before fusion into a single shared state, as motivated by the polarity structure of event measurements and learned event representations~\cite{Lichtsteiner2008DVS,Gehrig2019EndToEnd}.

\subsection{Transported spatial state}
\label{sec:spatial-state}
The retained memory $M_{k-1}\in\mathbb R^{B\times64\times260\times346}$ is transported in image coordinates by a bounded, learned backward sampling field:
\begin{align}
f_k(u)&=v_{\max}\Delta t_k\tanh\!\left(F_\theta([M_{k-1},O_k])(u)\right),\\
\widetilde M_{k-1}(u)&=\operatorname{Sample}(M_{k-1},u+f_k(u)).
\label{eq:transport}
\end{align}
A positive learned channelwise rate $r$ gives retention $\alpha_k=\exp(-r\Delta t_k)$. The state is updated by
\begin{equation}
\begin{split}
M_k={}&\alpha_k\odot\widetilde M_{k-1}\\
&+(1-\alpha_k)\odot S_k\odot\sigma(G_\theta(O_k))\odot\tanh(V_\theta(O_k)).
\end{split}
\label{eq:spatial-update}
\end{equation}
Transport and decay propagate existing context, including where the current packet has no events; support gating restricts new injection to observed pixels. Motion-compensated event reconstruction also uses temporal alignment of spatial evidence~\cite{Rebecq2018EMVS,Gallego2018Contrast}, but our displacement field transports learned features rather than reconstructing geometry from a supplied flow field. The retention coefficients compose with physical elapsed time, $\prod_k\exp(-r\Delta t_k)=\exp(-r\sum_k\Delta t_k)$, although the \emph{complete} state is not invariant to packet subdivision because transport and gating depend on intermediate observations.

\subsection{Multiscale event-conditioned depth head}
\label{sec:attention}
A four-level feed-forward pyramid is derived from $M_k$ at spatial factors $s\in\{1,2,4,8\}$, and a lightweight coarse-to-fine decoder yields spatial descriptors. Multi-resolution spatial features are a recurring component of monocular depth prediction~\cite{Ranftl2020TowardsRobust,Piccinelli2024UniDepth}; here the hierarchy is derived from a single event-updated state rather than storing additional recurrent memories. For each event, its 64-dimensional token is combined with normalized coordinates, signed polarity, relative packet time, and physical age $b_k-t_i$. At each scale, bilinear sampling extracts the centre descriptor and four neighboring offsets. The resulting 20 keys are aggregated by local cross-attention:
\begin{equation}
c_i=\sum_{j=1}^{20}\operatorname{softmax}_{j}\!\left(\frac{\langle W_Qq_i',W_Kk_{ij}\rangle}{\sqrt{d_a}}\right)W_Vk_{ij}.
\label{eq:local-attn}
\end{equation}
A residual event MLP predicts a scalar $\ell_i$ mapped into a bounded log-metric depth interval,
\begin{equation}
\hat d_i=\exp\!\left(\log d_{\min}+\sigma(\ell_i)(\log d_{\max}-\log d_{\min})\right).
\label{eq:depth-head}
\end{equation}
The outdoor checkpoint uses $d_{\min}=0.25$\,m and $d_{\max}=30$\,m. Depth outputs retain the input event indices after packet-wise processing. The query-to-key interaction count is $20N_k$ rather than all-pairs event attention; the spatial recurrence and pyramid have additional costs.

\subsection{Target-time geometric supervision}
\label{sec:supervision}
For event image coordinate $u_i=[x_i,y_i,1]^\top$ and predicted optical-axis depth $\hat d_i$, lift the event to $X_i=\hat d_iK^{-1}u_i$. Given camera-to-world poses $T(t)$, the same point in a reference camera at $t_r$ is
\begin{equation}
X_r=T(t_r)^{-1}T(t_i)X_i,\quad
(\tilde u_i^r,\tilde d_i^r)=(\pi(KX_r),[X_r]_z).
\label{eq:warp}
\end{equation}
The transform in Eq.~\eqref{eq:warp} accounts for camera motion alone and therefore assumes that the projected scene point remains stationary in world coordinates between $t_i$ and $t_r$. Independently moving objects violate this correspondence and may introduce erroneous depth supervision; visibility changes and occlusions further limit valid projections. To reduce these violations, regions containing complex independent object motion were cropped out during dataset preparation.

Translation is linearly interpolated and camera rotation interpolated by quaternion SLERP at event timestamps, consistent with the need for time-indexed camera geometry in event-based reconstruction and odometry~\cite{Mueggler2018ContinuousTime,Rebecq2018EMVS}. Valid projected depths are differentiably splatted to a target-time image $\widehat D_r$ with rendered support $\Omega_r$. The training loss compares this image with the reference depth $D_r^{\rm GT}$:
\begin{equation}
\mathcal L_{\rm SiLog}=\sqrt{\langle e^2\rangle_{\Omega_r}-\tfrac12\langle e\rangle_{\Omega_r}^{2}},\quad
e=\log D_r^{\rm GT}-\log\widehat D_r.
\label{eq:silog}
\end{equation}
Only pixels with valid rendered and reference depths contribute. The mean-residual penalty in Eq.~\eqref{eq:silog} retains sensitivity to metric depth. Pose annotations are used for geometric supervision and reference-time evaluation. Synchronized intensity images are \emph{training-only} inputs used exclusively for encoder distillation. The native inference input consists of events and recurrent memory.

\subsection{Cross-modal encoder initialization}
\label{sec:distillation}
Prior event-depth distillation studies exploit pretrained image representations to supply spatial cues where events are sparse~\cite{Bartolomei2025DepthAnyEvent,Zhu2025DepthAnyEventStream}. We use a frozen DINOv3 ViT-S/16 teacher to supply features from layers 2, 5, 8, and 11 at synchronized training images~\cite{Ch5Simeoni2025DINOv3}. Student spatial features are pooled to a $30\times40$ grid and projected to the teacher-token dimension. SmoothL1 feature regression, cosine alignment, and spatial-relation matching initialize the student representation before event-depth training. The teacher is discarded at inference, and the event encoder and newly initialized local-attention head are subsequently optimized for target-time metric-depth supervision.
\section{Experiments}
\label{sec:experiments}
\subsection{Evaluation protocol}
\label{sec:exp-protocol}
\begin{table*}[t]
    \centering
    \small
    \setlength{\tabcolsep}{2.5pt}\renewcommand{\arraystretch}{0.85}
    \setlength{\aboverulesep}{0pt} \setlength{\belowrulesep}{0pt}
    \caption{Per-sequence metric-depth accuracy on the MVSEC outdoor recordings.}
    \label{tab:ch3:depth-per-seq-mvsec}
    \label{tab:attention_depth_main_metrics}
    \resizebox{0.9\linewidth}{!}{%
        \begin{tabular}{@{}llccccccc@{}}
            \toprule
            Sequence & Method & AbsRel $\downarrow$ & SqRel $\downarrow$ & RMSE (m) $\downarrow$ & MAE (m) $\downarrow$ & $\delta_1$ (\%) $\uparrow$ & $\delta_2$ (\%) $\uparrow$ & $\delta_3$ (\%) $\uparrow$ \\
            \midrule
            \multirow{5}{*}{Day 1} & E2Depth & 0.729 & 9.195 & 8.182 & 5.612 & 36.99 & 60.86 & 75.36 \\
            & EReFormer & 0.702 & 7.678 & 8.646 & 6.272 & 31.53 & 57.25 & 74.18 \\
            & DepthAnyEvent-R & 0.599 & 6.097 & 10.536 & 7.984 & 8.16 & 18.32 & 27.86 \\
            & ScaleEvent & \underline{0.357} & \underline{3.174} & \underline{6.353} & \underline{3.964} & \textbf{55.71} & \textbf{78.23} & \underline{90.11} \\
            & EventSSM (Ours) & \textbf{0.286} & \textbf{1.413} & \textbf{4.745} & \textbf{3.152} & \underline{46.77} & \underline{76.80} & \textbf{91.39} \\
            \midrule
            \multirow{5}{*}{Night 1} & E2Depth & 1.020 & 16.868 & 9.028 & 6.166 & 36.28 & 60.10 & 73.29 \\
            & EReFormer & 0.598 & 6.870 & 7.690 & 5.232 & 42.94 & 67.06 & 80.10 \\
            & DepthAnyEvent-R & 0.633 & 6.962 & 11.367 & 8.848 & 6.62 & 16.53 & 24.85 \\
            & ScaleEvent & \underline{0.318} & \underline{2.307} & \underline{5.889} & \underline{3.898} & \underline{45.86} & \underline{72.43} & \underline{87.70} \\
            & EventSSM (Ours) & \textbf{0.304} & \textbf{1.664} & \textbf{5.001} & \textbf{3.446} & \textbf{46.56} & \textbf{77.06} & \textbf{91.41} \\
            \midrule
            \multirow{5}{*}{Night 2} & E2Depth & 0.712 & 10.754 & 8.171 & 4.887 & 40.12 & 65.69 & 80.19 \\
            & EReFormer & 0.354 & \underline{2.104} & \underline{5.131} & 3.523 & \underline{44.83} & \underline{72.78} & 89.09 \\
            & DepthAnyEvent-R & 0.613 & 5.728 & 9.771 & 7.523 & 6.01 & 15.31 & 25.82 \\
            & ScaleEvent & \underline{0.314} & 2.243 & 5.656 & \underline{3.453} & \textbf{50.09} & \textbf{75.83} & \textbf{90.38} \\
            & EventSSM (Ours) & \textbf{0.310} & \textbf{1.470} & \textbf{4.597} & \textbf{3.217} & 40.16 & 72.05 & \underline{89.75} \\
            \midrule
            \multirow{5}{*}{Night 3} & E2Depth & 0.797 & 13.328 & 8.803 & 5.315 & 37.04 & 64.28 & 79.58 \\
            & EReFormer & 0.319 & \underline{1.764} & \underline{5.030} & \underline{3.435} & \underline{46.07} & \underline{74.31} & \underline{90.50} \\
            & DepthAnyEvent-R & 0.621 & 6.098 & 10.217 & 7.902 & 6.47 & 16.12 & 25.20 \\
            & ScaleEvent & \underline{0.303} & 2.153 & 5.772 & 3.566 & \textbf{49.15} & \textbf{75.26} & 89.89 \\
            & EventSSM (Ours) & \textbf{0.299} & \textbf{1.467} & \textbf{4.642} & \textbf{3.216} & 43.26 & \underline{74.31} & \textbf{90.84} \\
                    \bottomrule
        \end{tabular}%
        }
\end{table*}

We evaluate on MVSEC~\cite{Zhu2018MVSEC}, training on \texttt{outdoor\_day2} and selecting checkpoints using native AbsRel on \texttt{outdoor\_day1}. The three outdoor-night sequences form the held-out evaluation set; Day~1 is also the checkpoint-selection sequence, and its results are interpreted accordingly. An eligible approximately 50\,ms reference interval contains at least 10,000 natural events and has camera-pose annotations within 5\,ms of both depth endpoints. We select 7,500 events while retaining their chronology and divide each interval into six non-overlapping equal-duration packets; packet event counts need not be equal. The spatial memory is initialised at the beginning of each independently sampled training or checkpoint-evaluation interval. Training uses 15 epochs, batch size four, Adam optimisation and a validation-driven learning-rate schedule.

Because target-time supervision uses camera motion to relate event observations to the annotated view, the preparation protocol excludes image regions containing complex independently moving objects through cropping. These regions would not generally obey the static-world correspondence assumed by the camera-only warp. The reported evaluation uses the retained image regions and matched spatial scoring masks.

Evaluation uses matched reference timestamps, windows and spatial scoring masks across the compared depth methods. The data preprocessing will be released to the community to support future comparisons. EventSSM estimates are transported to the annotated viewpoint using the camera trajectory and rendered onto the reference-time image, whereas dense baseline depth maps are sampled at the same scored locations. EReFormer's official test code crops its input to a fixed $224\times224$ window %
so its prediction covers only that window. To ensure fairness of comparison, the scoring masks are shared across every method in Tables~\ref{tab:ch3:depth-per-seq-mvsec} and~\ref{tab:ch3:depth-range-mae}. AbsRel and SqRel measure the relative and squared-relative residuals. RMSE and MAE measure errors in metres. The threshold accuracies $\delta_j$ give the percentage of estimates satisfying $\max(\hat d/d,d/\hat d)<1.25^j$ for $j\in\{1,2,3\}$. For MAE@$r$, the mean absolute error is restricted to valid reference depths no greater than $r$ metres, with $r\in\{10,20,30\}$. Missing predictions are not imputed as correct estimates. The fixed evaluation windows and masks provide direct numerical comparisons within the stated protocol.

\subsection{MVSEC depth comparison}
\label{sec:exp-depth}
\begin{table}[t]
\centering
\caption{Mean absolute depth error (metres) within the specified ground-truth depth limits on MVSEC.}
\label{tab:ch3:depth-range-mae}
\scriptsize
\setlength{\tabcolsep}{2.6pt}
\renewcommand{\arraystretch}{1.07}
\begin{tabular}{@{}llrrr@{}}
\toprule
Sequence & Method & MAE@10 $\downarrow$ & MAE@20 $\downarrow$ & MAE@30 $\downarrow$ \\
\midrule
Day 1 & E2Depth & 4.938 & 5.338 & 5.612 \\
 & EReFormer & 5.232 & 6.229 & 6.272 \\
 & DepthAnyEvent-R & 2.809 & 5.764 & 7.984 \\
 & ScaleEvent & \underline{2.200} & \underline{3.431} & \underline{3.964} \\
 & EventSSM (Ours) & \textbf{1.544} & \textbf{2.300} & \textbf{3.152} \\
\midrule
Night 1 & E2Depth & 7.136 & 6.109 & 6.166 \\
 & EReFormer & 4.541 & 4.825 & 5.232 \\
 & DepthAnyEvent-R & 2.937 & 6.368 & 8.848 \\
 & ScaleEvent & \textbf{1.704} & \underline{2.850} & \underline{3.898} \\
 & EventSSM (Ours) & \underline{1.782} & \textbf{2.526} & \textbf{3.446} \\
\midrule
Night 2 & E2Depth & 4.688 & 4.631 & 4.887 \\
 & EReFormer & 2.112 & 2.893 & 3.523 \\
 & DepthAnyEvent-R & 3.081 & 5.982 & 7.523 \\
 & ScaleEvent & \textbf{1.682} & \underline{2.831} & \underline{3.453} \\
 & EventSSM (Ours) & \underline{1.797} & \textbf{2.582} & \textbf{3.217} \\
\midrule
Night 3 & E2Depth & 5.312 & 5.048 & 5.315 \\
 & EReFormer & 1.782 & \underline{2.664} & \underline{3.435} \\
 & DepthAnyEvent-R & 3.075 & 6.167 & 7.902 \\
 & ScaleEvent & \textbf{1.570} & 2.802 & 3.566 \\
 & EventSSM (Ours) & \underline{1.760} & \textbf{2.547} & \textbf{3.216} \\
\bottomrule
\end{tabular}
\end{table}

Table~\ref{tab:ch3:depth-per-seq-mvsec} compares EventSSM with four established event-depth estimators using their published codebases on common reference-time geometry. EventSSM leads all four sequences in AbsRel, SqRel, RMSE and MAE, improving both relative depth consistency and absolute geometric error. Against the strongest competing method for each measure, the reductions on Day~1 reach 19.9\% in AbsRel, 55.5\% in SqRel, 25.3\% in RMSE and 20.5\% in MAE. The advantage persists on the held-out night sequences: SqRel falls by 16.8--30.1\% and RMSE by 7.7--15.1\% relative to the best corresponding baseline. These improvements across different lighting conditions support the value of accumulating event evidence in a persistent spatial representation.

\providecommand{\QualRowDay}{outdoor\\day1}
\providecommand{\QualRowNightOne}{outdoor\\night1}
\providecommand{\QualRowNightTwo}{outdoor\\night2}
\providecommand{\QualRowNightThree}{outdoor\\night3}
\renewcommand{\QualRowDay}{outdoor\\day1}
\renewcommand{\QualRowNightOne}{outdoor\\night1}
\renewcommand{\QualRowNightTwo}{outdoor\\night2}
\renewcommand{\QualRowNightThree}{outdoor\\night3}

\begin{figure*}[t]
    \centering
    \begingroup
    \setlength{\tabcolsep}{0pt}
    \renewcommand{\arraystretch}{0.6}
    \setlength{\fboxsep}{0pt}
    \setlength{\fboxrule}{0.4pt}
    \newcommand{\qualw}{0.127\textwidth}
    \newcommand{\qualgap}{\hspace{3.6pt}}%
    \newcommand{\qualsep}{\hspace{7pt}}%
    \newcommand{\colhead}[1]{\makebox[\dimexpr\qualw+2\fboxrule\relax][c]{\bfseries\sffamily\scriptsize\shortstack[c]{#1}}}
    \newcommand{\rowlabel}[1]{\makebox[0.03\textwidth][c]{\rotatebox[origin=c]{90}{\bfseries\sffamily\scriptsize\shortstack[c]{#1}}}}
    \newcommand{\qualimg}[1]{\raisebox{-0.5\height}{\fcolorbox{black!30}{white}{\IfFileExists{#1}%
        {\includegraphics[width=\qualw]{#1}}%
        {\parbox[c][0.75\qualw][c]{\qualw}{\centering\scriptsize Panel\\pending}}}}}
    \newcommand{\winimg}[1]{\begin{tikzpicture}
        \node[inner sep=0pt, anchor=south west] (p) at (0,0) {\includegraphics[width=\qualw]{#1}};
        \begin{scope}[x={(p.south east)}, y={(p.north west)}]
            \draw[overlay, black!75, line width=0.5pt, dash pattern=on 1.6pt off 1.2pt] (121/346, 3/260) rectangle (345/346, 227/260);
        \end{scope}\end{tikzpicture}}
    \newcommand{\qualimgwin}[1]{\raisebox{-0.5\height}{\fcolorbox{black!30}{white}{\IfFileExists{#1}%
        {\winimg{#1}}%
        {\parbox[c][0.75\qualw][c]{\qualw}{\centering\scriptsize Panel\\pending}}}}}
    \begin{tabular}{@{}c@{\hspace{3pt}}c@{\qualgap}c@{\qualsep}c@{\qualsep}c@{\qualgap}c@{\qualgap}c@{\qualgap}c@{}}
        \makebox[0.03\textwidth]{}
        & \colhead{Event\\image}
        & \colhead{Ground\\truth}
        & \colhead{EventSSM\\(Ours)}
        & \colhead{E2Depth\\\phantom{x}}
        & \colhead{EReFormer\\\phantom{x}}
        & \colhead{DepthAny-\\Event-R}
        & \colhead{ScaleEvent\\\phantom{x}} \\ \noalign{\vskip 2pt}
        \rowlabel{\QualRowDay} &
        \qualimg{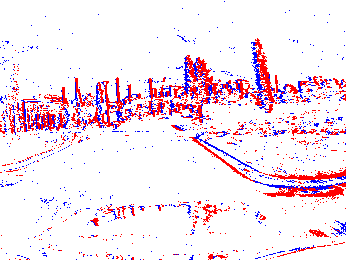} &
        \qualimg{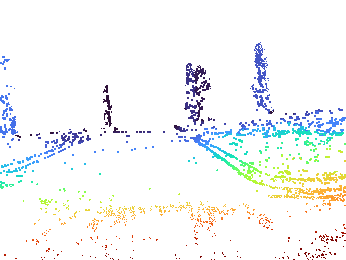} &
        \qualimg{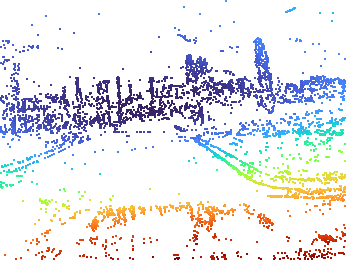} &
        \qualimg{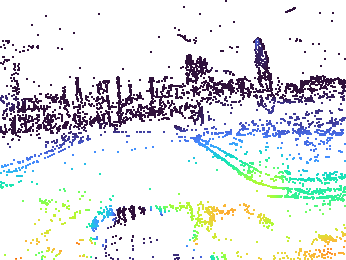} &
        \qualimgwin{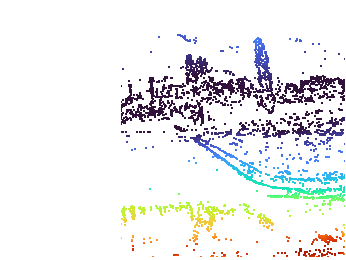} &
        \qualimg{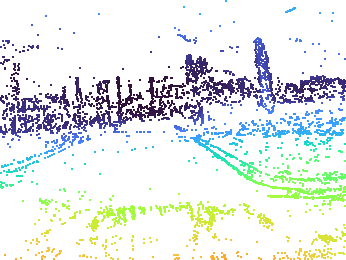} &
        \qualimg{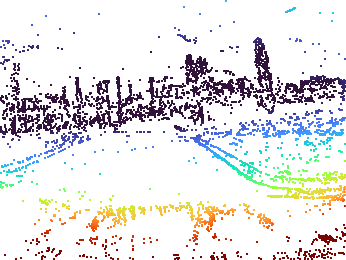} \\ \noalign{\vskip 4pt}
        \rowlabel{\QualRowNightOne} &
        \qualimg{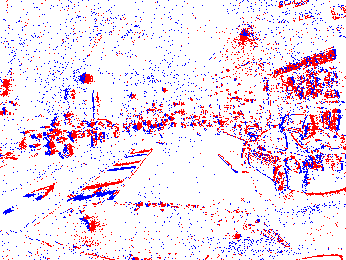} &
        \qualimg{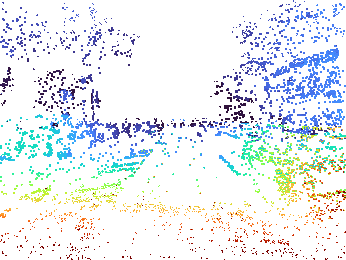} &
        \qualimg{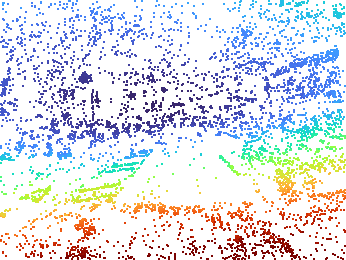} &
        \qualimg{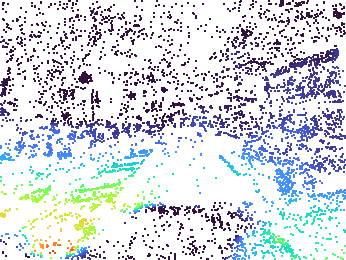} &
        \qualimgwin{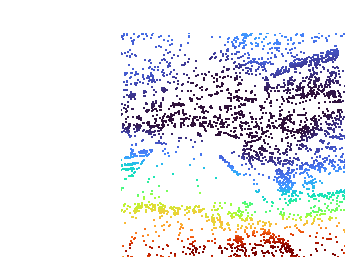} &
        \qualimg{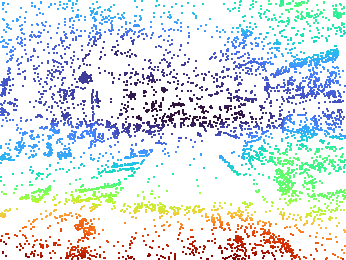} &
        \qualimg{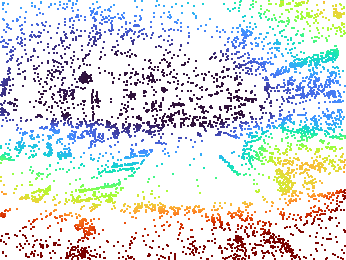} \\ \noalign{\vskip 4pt}
        \rowlabel{\QualRowNightTwo} &
        \qualimg{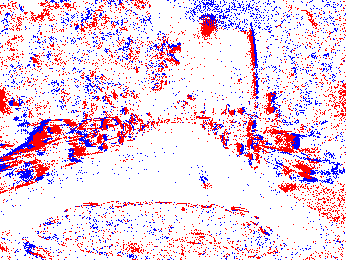} &
        \qualimg{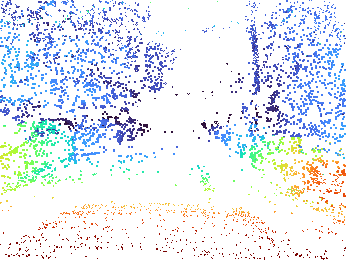} &
        \qualimg{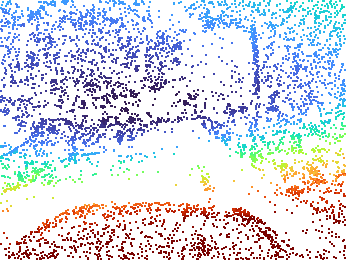} &
        \qualimg{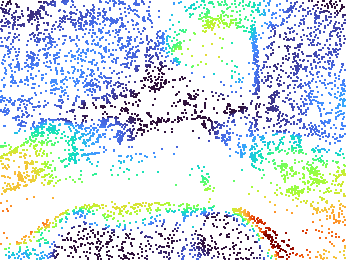} &
        \qualimgwin{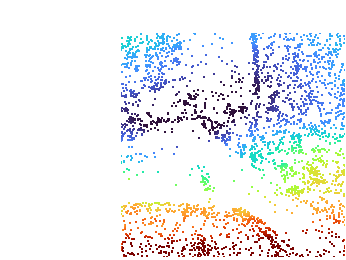} &
        \qualimg{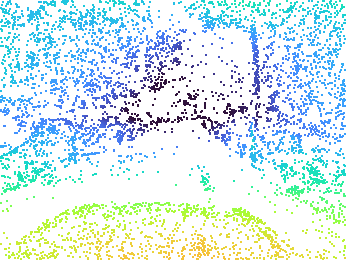} &
        \qualimg{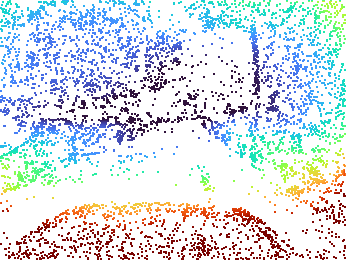} \\ \noalign{\vskip 4pt}
        \rowlabel{\QualRowNightThree} &
        \qualimg{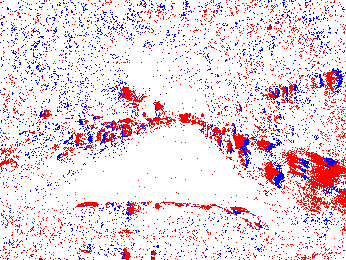} &
        \qualimg{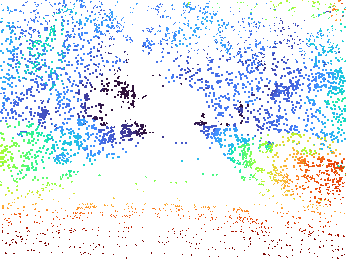} &
        \qualimg{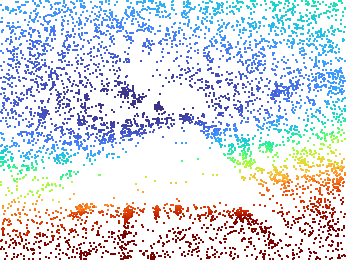} &
        \qualimg{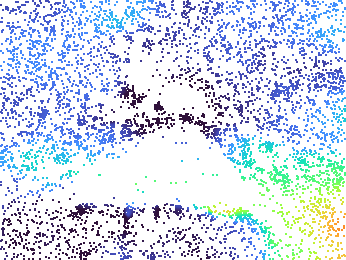} &
        \qualimgwin{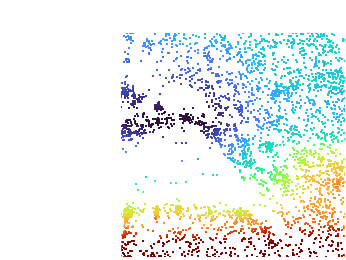} &
        \qualimg{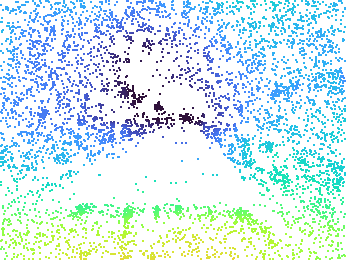} &
        \qualimg{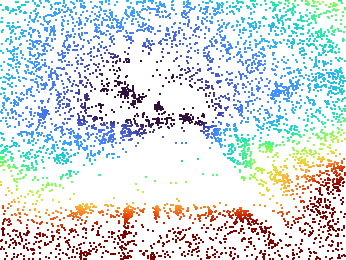}
    \end{tabular}
    \endgroup
    \caption{Qualitative metric depth on MVSEC outdoor sequences. EventSSM predicts depth for individual events, shown rendered at the reference time; for a like-for-like view, the ground truth and the baselines' dense maps are shown only at those event pixels. EReFormer predicts only inside the fixed lower-right $224\times224$ window of its official test code (dashed), shown at its position in the frame. Colours encode inverse depth with a shared range per row (red is near). White marks pixels without a value.}
    \label{fig:ch3:continuous-depth-qualitative}%
\end{figure*}

The threshold measures provide a complementary view of prediction quality. EventSSM achieves the strongest accuracy at all three tolerances on Night~1, and the highest $\delta_3$ on Day~1 and Night~3. ScaleEvent leads at some tighter thresholds, while EventSSM consistently lowers the aggregate depth errors. The combined results therefore favour EventSSM for reducing the magnitude of geometric residuals.

The distance-restricted comparison in Table~\ref{tab:ch3:depth-range-mae} further identifies where the gains arise. EventSSM leads MAE@20 and MAE@30 on every evaluated sequence, reducing the best competing errors by 4.4--33.0\% and 6.4--20.5\%, respectively. At the nearest 10\,m cutoff, its Day~1 error is 29.8\% below the strongest alternative, while ScaleEvent leads on the three night sequences. The broader-range consistency is significant: predictions remain accurate as evaluation includes more distant geometry, despite the irregular spatial sampling of asynchronous events.

\begin{table}[H] 
    \centering 
    \caption{Effect of cross-modal encoder initialization on depth estimation.} 
    \label{tab:ch3:distillation-ablation} 
    \begin{tabular*}{\linewidth}{@{\extracolsep{\fill}}lc@{}} \toprule Variant & Depth Error (m) $\downarrow$ \\ 
        \midrule Without distillation & \underline{2.724} \\ 
        With distillation & \textbf{2.331} \\ 
        \bottomrule 
    \end{tabular*} 
\end{table}

Figure~\ref{fig:ch3:continuous-depth-qualitative} complements these aggregate findings with reference-time reconstructions under daylight and night conditions. Event-supported predictions preserve visible scene structure despite highly nonuniform activity across textured boundaries and less active surfaces. The qualitative comparison illustrates how a spatially contextualised event representation supports depth recovery beyond a purely local interpretation of each brightness change.

\subsection{Architecture and initialization ablations}
\label{sec:exp-ablation}
The cross-modal initialization study (Table~\ref{tab:ch3:distillation-ablation}) shows that transferring image-trained spatial features reduces downstream depth error by 14.4\%. This establishes the benefit of a stronger spatial prior for the event encoder, while inference remains entirely event-based and does not depend on the image teacher. The gain complements, rather than replaces, the causal memory and event-conditioned prediction mechanism.

\begin{table}[H]
    \centering
    \caption[Temporal-backbone ablation]{Depth accuracy, model size and inference time for undistilled temporal-state encoders. All packet-level inference measurements were obtained on one NVIDIA GeForce RTX 2080 Ti, in FP32 with batch size one}
    \label{tab:ch3:spatial-core-ablation}
    \begin{adjustbox}{max width=\linewidth}
        \begin{tabular}{@{}lrrr@{}}
            \toprule
            Temporal backbone & \shortstack{Params.\\(k) $\downarrow$} & \shortstack{Depth error\\(m) $\downarrow$} & \shortstack{Packet time\\(ms) $\downarrow$} \\
            \midrule
            Spatial RNN & \underline{73.6} & 3.566 & \underline{21.847} \\
            Spatial LSTM & 98.4 & \underline{3.543} & 23.433 \\
            bi-Mamba & 79.2 & 3.596 & 27.423 \\
            EventSSM (ours) & \textbf{69.5} & \textbf{3.507} & \textbf{5.519} \\
            \bottomrule
        \end{tabular}
    \end{adjustbox}
\end{table}

The undistilled temporal-core comparison (Table~\ref{tab:ch3:spatial-core-ablation}) isolates the choice of recurrent update under a common 64-channel configuration. EventSSM attains 1.0--2.5\% lower depth error than the RNN, LSTM and bi-Mamba alternatives while using 5.6--29.4\% fewer parameters. Its most pronounced advantage is computational: packet inference is approximately 4.0--5.0$\times$ faster than those recurrent alternatives on the same GPU. This combination of comparable accuracy and substantially lower update latency motivates the proposed spatial-state transition for fine-grained event processing. The initialization and temporal-core studies use distinct training conditions and are interpreted independently.

\subsection{Packet-count sensitivity}
\label{sec:exp-packets}
\begin{figure}[t]
    \centering
    \includegraphics[width=1.05\linewidth]{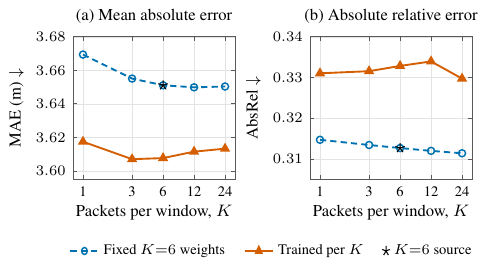}
    \caption[Day1 packet-count accuracy for per-event depth]{Per-event depth on \texttt{outdoor\_day1} versus packets per window $K$: (a) native MAE and (b) native AbsRel. Dashed curves retain the $K=6$ source weights; solid curves use separate $K$-specific full-model training with a 20-epoch budget and guarded Day1 AbsRel selection. Stars mark the source at $K=6$.}
    \label{fig:ch3:packet-count-day1}
\end{figure}

Figure~\ref{fig:ch3:packet-count-day1} tests whether the representation remains effective as the same annotated event interval is divided into different numbers of causal packets. The common Day~1 validation population and scoring intersection make the accuracy trends directly comparable within this ablation. With the original $K=6$ weights retained, relative error changes little around the three- to six-packet settings, indicating that the learned memory tolerates moderate changes in update frequency. Independently trained packet-specific configurations improve MAE but not AbsRel uniformly, and increasing $K$ does not produce monotonic accuracy gains. Thus, finer temporal subdivision is an operational design choice rather than a guarantee of better reference-time error.

We retain $K=6$ as a practical balance between temporal responsiveness and repeated state-update cost. Dividing an approximately 50\,ms annotated interval into six chronological packets permits an event-associated depth readout after roughly 8.3\,ms of observations, instead of waiting for the next annotated endpoint. The packetisation is used during supervised training to expose causal intermediate states; a deployed model can continue updating its memory as new packets arrive without requiring a ground-truth-aligned window. These findings concern accuracy on the shared 3,066-window validation subset, whose intersection covers 16.98\% of eligible ground-truth pixels, while inference cost is evaluated separately below.

\subsection{Component ablations}
\label{sec:exp-components}
\begin{table}[tb]
    \centering
    \caption{Component ablations. Mean absolute depth error, averaged over evaluation windows. Values are the mean $\pm$ half-range over two training seeds; each variant changes one component of the full model.}
    \label{tab:ch3:component-ablation}
    \begin{adjustbox}{max width=\linewidth}
        \begin{tabular}{@{}lrcc@{}}
            \toprule
            Variant & \shortstack{Params.\\(k)} & \shortstack{Day~1\\MAE (m) $\downarrow$} & \shortstack{Nights\\MAE (m) $\downarrow$} \\
            \midrule
            EventSSM (full) & 69.5 & 3.553 $\pm$ 0.004 & 3.534 $\pm$ 0.002 \\
            \midrule
            \multicolumn{4}{@{}l}{\emph{Spatial state (Sec.~\ref{sec:spatial-state})}} \\
            No transport & 67.0 & 3.564 $\pm$ 0.049 & 3.541 $\pm$ 0.007 \\
            Per-update retention & 69.5 & 3.616 $\pm$ 0.016 & 3.620 $\pm$ 0.039 \\
            No forgetting & 69.4 & 3.540 $\pm$ 0.021 & 3.521 $\pm$ 0.007 \\
            No input gate & 65.3 & 3.621 $\pm$ 0.085 & 3.537 $\pm$ 0.036 \\
            No support mask & 69.5 & 3.597 $\pm$ 0.033 & 3.566 $\pm$ 0.031 \\
            \midrule
            \multicolumn{4}{@{}l}{\emph{Depth head (Sec.~\ref{sec:attention})}} \\
            Average-pooled scales & 49.8 & 3.555 $\pm$ 0.005 & 3.511 $\pm$ 0.008 \\
            Centre-sample readout & 70.5 & 3.556 $\pm$ 0.086 & 3.578 $\pm$ 0.072 \\
            \midrule
            \multicolumn{4}{@{}l}{\emph{Supervision (Sec.~\ref{sec:supervision})}} \\
            No target-time warp & 69.5 & 3.535 $\pm$ 0.002 & 3.535 $\pm$ 0.006 \\
            \bottomrule
        \end{tabular}
    \end{adjustbox}
\end{table}

Table~\ref{tab:ch3:component-ablation} changes one component of the spatial state (Sec.~\ref{sec:spatial-state}), the depth head (Sec.~\ref{sec:attention}) or the training supervision (Sec.~\ref{sec:supervision}) at a time. \emph{No transport} sets $\widetilde M_{k-1}=M_{k-1}$. \emph{Per-update retention} replaces $\exp(-r\Delta t_k)$ by a learned channelwise constant that does not depend on $\Delta t_k$. \emph{No forgetting} replaces $\alpha_k\odot\widetilde M_{k-1}$ in Eq.~\eqref{eq:spatial-update} by $\widetilde M_{k-1}$, with the rate fixed at its initial value. \emph{No input gate} and \emph{no support mask} set $\sigma(G_\theta(O_k))\equiv1$ and $S_k\equiv1$, respectively. The latter also writes new evidence at pixels without events. \emph{Average-pooled scales} replaces the learned pyramid by average pooling of $M_k$ at the same factors, and \emph{centre-sample readout} passes the four concatenated centre descriptors to the event MLP instead of attending over the 20 keys of Eq.~\eqref{eq:local-attn}. \emph{No target-time warp} sets $T(t_i)=T(t_r)$ in Eq.~\eqref{eq:warp} during training only, so each predicted depth is compared with the reference depth at its own pixel.

The spatial-state and pooled-scale variants are distilled with their change in place (Sec.~\ref{sec:distillation}). The other variants reuse the full model's encoder. Event heads are trained with the protocol of Sec.~\ref{sec:exp-protocol}, including selection by Day~1 AbsRel, the full model is retrained identically, and every row uses the same two seeds. Errors are scored on each model's own rendered support over the full frame, not on the masks shared with the baselines, so they are compared only within Table~\ref{tab:ch3:component-ablation}.

The full-model results agree across seeds to within 0.01\,m, whereas the seeds of some variants differ by up to 0.17\,m. Removing transport, forgetting or the input gate, or using centre samples, changes the mean error by at most 0.07\,m on Day~1 and 0.05\,m at night. The results across seeds for each of these variants include errors at or below the full model on Day~1, and their AbsRel stays within 0.010 of the full model. Removing the support mask raises the error by 0.045\,m (1.3\%) on Day~1 and 0.032\,m (0.9\%) at night. Its errors consistently exceed the full model across seeds on Day~1, whereas the night results include a match. Under the reset-window protocol the memory spans only six packets of about 8.3\,ms, so transported and retained context has little time to accumulate.

Per-update retention is the only change whose error range across seeds lies entirely above that of the full model, on Day~1 and at night: the error rises by 0.064\,m (1.8\%) and 0.087\,m (2.5\%), and the night increase remains after per-frame median scaling (0.093\,m). All packets have equal duration, so both forms can express the same retention. The gap therefore reflects how retention is parameterised and learned.

Average-pooled scales remove 19.7k parameters (28\%) without increasing the error ($+0.002$\,m on Day~1, $-0.022$\,m at night), and the results agree across seeds to within 0.016\,m. Fixed pooling of the state thus supplies the multiscale context, making the 49.8k-parameter variant an equally accurate, more compact alternative in this protocol.

Training without the target-time warp matches the full model: at night the error changes by $+0.002$\,m, with the results across seeds on either side of the full model, and on Day~1, the selection sequence, it is 0.018\,m (0.5\%) lower. AbsRel differs by at most 0.002, and after median scaling the error is 0.006--0.009\,m higher. Over about 50\,ms the camera moves little, so warping the supervision does not change accuracy here. Selection and evaluation use Eq.~\eqref{eq:warp} for every row.

\subsection{Inference-only computation}
\label{sec:exp-runtime}
\begin{table}[t]
    \centering
    \caption[Inference runtime comparison]{Inference time and model complexity on an NVIDIA GeForce RTX 2080 Ti (FP32, batch size one).}
    \label{tab:ch3:depth-baseline-runtime-same-node}
    \begin{adjustbox}{max width=\linewidth}
        \begin{tabular}{@{}lrrrrr@{}}
            \toprule
            Method & Params. (M) $\downarrow$ & Mean (ms) $\downarrow$ & Calls/s $\uparrow$ & RTF $\downarrow$ & GFLOPs$_{\mathrm{ref}}$ $\downarrow$ \\
            \midrule
            E2Depth & \underline{10.7104} & 11.88 & 84.16 & \underline{0.238} & 84.393 \\
            EReFormer & 89.5291 & 44.86 & 22.29 & 0.897 & 39.061 \\
            DepthAnyEvent-R & 26.1135 & \underline{10.59} & \underline{94.47} & \textbf{0.212} & \underline{38.235} \\
            ScaleEvent & 24.3297 & 15.76 & 63.47 & 0.315 & 99.097 \\
            \midrule
            EventSSM (ours; packet) & \textbf{0.0695} & \textbf{5.58} & \textbf{179.11} & 0.670 & \textbf{5.541} \\
            \bottomrule
        \end{tabular}%
    \end{adjustbox}
    \vspace{-0.3cm}
\end{table}

Table~\ref{tab:ch3:depth-baseline-runtime-same-node} highlights the computational advantage of the event-conditioned architecture. EventSSM has an approximately 154$\times$ smaller parameter count than the smallest compared full-window baseline, and its cost is about 6.9$\times$ below the next-lowest baseline. Its packet call has the shortest measured inference latency and the highest call rate, supporting more frequent depth updates from a compact model.

The temporal support of each call matters when interpreting these figures. EventSSM produces six packet-boundary predictions over an approximately 50\,ms interval, whereas the baselines make one complete-window prediction. The real-time factor (RTF) divides the mean inference time of a call by the sensor time that call covers, approximately 50\,ms for a baseline window and 8.33\,ms for an EventSSM packet. Values below one mean that inference keeps pace with the sensor. The measured packet latency is below its approximately 8.33\,ms observation span, giving an inference-only RTF of 0.670 and leaving computational headroom for successive updates. Its RTF is consequently higher than those of three window-based baselines because the model performs more readouts per unit of sensor time. The benefit is a finer prediction cadence rather than a claim of lower total window-normalised execution cost. These measurements use FP32, batch size one, and exclude input preparation, host transfers, rendering and scoring.

\section{Conclusion}
\label{sec:conclusion}
We introduced EventSSM, a causal spatial-state framework that combines persistent motion-transported memory with local event-conditioned metric-depth prediction. Target-time rendering relates event-time predictions to sparse reference annotations for supervision and evaluation without restricting predictions to frame timestamps. Across the MVSEC outdoor comparisons, EventSSM consistently improves the principal relative and metric depth errors, with particularly strong squared-relative and wider-range accuracy. Ablations establish the benefits of cross-modal spatial initialization, the exponential retention parametization and the efficiency of the recurrent update, while packet-count experiments clarify the choice of temporal resolution. The resulting compact model supports depth readouts at successive short packet boundaries, advancing event-native geometric perception through a combination of accuracy, spatial context and fine-grained temporal responsiveness.

{
    \small
    \bibliographystyle{ieeenat_fullname}
    \bibliography{main}
}

\clearpage
\setcounter{page}{1}
\maketitlesupplementary
\appendix

\section{Model details for Sec.~\ref{sec:method}}
\label{sec:supp-method}
\begin{table*}[t]
    \centering
    \footnotesize
    \setlength{\tabcolsep}{3pt}
    \caption{Network and training settings of EventSSM (69.5k parameters). Feature maps are channels\,$\times$\,height\,$\times$\,width at the native $260\times346$ resolution; $N_k$ is the number of events in packet $\mathcal E_k$.}
    \label{tab:supp:settings}
    \begin{minipage}[t]{0.555\textwidth}
        \vspace{0pt}
        \begin{tabular}{@{}l@{\hspace{4pt}}p{0.585\linewidth}@{\hspace{4pt}}l@{}}
            \toprule
            Stage & Operation & Output \\
            \midrule
            Event token & MLP of normalized $(x,y)$, packet time $\tau_i$ and polarity; polarity embedding and residual; LayerNorm & $N_k\times64$ \\
            Observation & Per-polarity token means and $\log(1+n_k^{\pm})$ (130 ch.); $1\times1$, GELU, $3\times3$ depthwise & $64\times H\times W$ \\
            Transport & Backward field, $\|f_k\|_\infty\le v_{\max}\Delta t_k$, $v_{\max}=800$\,px\,s$^{-1}$ & $64\times H\times W$ \\
            State update & Eq.~\eqref{eq:spatial-update}; rates initialized to a 0.1\,s time constant & $64\times H\times W$ \\
            Pyramid & Average pool by $s\in\{1,2,4,8\}$, GroupNorm, $3\times3$ depthwise, GELU, $1\times1$ & 64 per scale \\
            Decoder & $1\times1$ to 32 ch.; coarse-to-fine bilinear upsampling, addition, residual block & 32 per scale \\
            Event query & Token, $(x,y)$, polarity, $\tau_i$, $\log(1+\mathrm{age}_{\mathrm{ms}})$ & $N_k\times32$ \\
            Attention & Eq.~\eqref{eq:local-attn}: event position and its 4 neighboring cells on 4 scales (20 keys) & $N_k\times32$ \\
            Readout & Residual MLP; Eq.~\eqref{eq:depth-head}, $d_{\min}=0.25$\,m, $d_{\max}=30$\,m & $N_k\times1$ \\
            \bottomrule
        \end{tabular}
    \end{minipage}\hfill
    \begin{minipage}[t]{0.425\textwidth}
        \vspace{0pt}
        \begin{tabular}{@{}l@{\hspace{4pt}}p{0.73\linewidth}@{}}
            \toprule
            Setting & Value \\
            \midrule
            \multicolumn{2}{@{}l}{\emph{Stage 1: encoder distillation (Sec.~\ref{sec:distillation})}} \\
            Teacher & DINOv3 ViT-S/16, frozen; layers 2, 5, 8, 11 \\
            Targets & Pyramid pooled to $30\times40$, $64\to384$ ch.; SmoothL1, cosine, spatial relation; 20 epochs \\
            \multicolumn{2}{@{}l}{\emph{Stage 2: depth training (Sec.~\ref{sec:supervision})}} \\
            Objective & $\mathcal L_{\rm SiLog}$ (Eq.~\eqref{eq:silog}); new attention head \\
            Optimizer & Adam, weight decay $10^{-2}$, batch 4, 15 epochs \\
            Learning rate & Encoder $10^{-6}$, head $2\times10^{-4}$; plateau on val.\ AbsRel (factor 0.5, patience 4, threshold $10^{-3}$, cooldown 1, min.\ $10^{-7}$) \\
            Other & Gradient norm 1; FP32 \\
            \multicolumn{2}{@{}l}{\emph{Input windows}} \\
            Intervals & \texttt{outdoor\_day2}; $\geq$10,000 events; poses within 5\,ms of both ends \\
            Events & 7,500 per interval (Sec.~\ref{sec:supp-sampling}); $K=6$ packets \\
            Depth range & Supervision 0.25--30\,m; evaluation 2--30\,m \\
            \bottomrule
        \end{tabular}
    \end{minipage}
\end{table*}

Table~\ref{tab:supp:settings} lists the network of Fig.~\ref{fig:ch3:continuous-depth-architecture} layer by layer for one packet, with the output size of each stage, and the settings of both training stages.

\paragraph{Packet interface (Sec.~\ref{sec:method-formulation}, Eq.~\eqref{eq:state-interface}).}
Packet $\mathcal E_k$ spans $[b_{k-1},b_k]$, so $\Delta t_k=b_k-b_{k-1}$. The depths $\hat{\boldsymbol d}_k$ of a packet become available at its endpoint $b_k$, after its last event has arrived, so the packet endpoint is the earliest output time of every event in $\mathcal E_k$. The validity mask of Sec.~\ref{sec:method-formulation} also excludes padded queries from the loss and from evaluation.

\paragraph{Polarity-aware observations (Sec.~\ref{sec:tokenizer}).}
Each event token $q_i$ is a linear--ReLU--linear projection of $\bigl(2x_i/(W-1)-1,\,2y_i/(H-1)-1,\,\tau_i,\,p_i\bigr)$, with $W\times H=346\times260$ and within-packet time $\tau_i=(t_i-b_{k-1})/\Delta t_k\in[0,1]$, to which a polarity embedding and a polarity-specific residual embedding are added before layer normalization. In the spatial stem, a polarity group without events at a pixel contributes a zero mean and a zero log-count, while the other polarity contributes normally, and padded events are removed before the counts and means are formed.

\paragraph{Transport and retention (Sec.~\ref{sec:spatial-state}, Eqs.~\eqref{eq:transport} and~\eqref{eq:spatial-update}).}
In Eq.~\eqref{eq:transport}, $\mathrm{Sample}$ is bilinear resampling and the last layer of $F_\theta$ is initialized to zero, so training starts from identity transport. The checkpoint uses $v_{\max}=800$\,px\,s$^{-1}$, so the displacement of the state per packet is bounded coordinatewise by $v_{\max}\Delta t_k$, about 6.7\,px for an 8.3\,ms packet. The channelwise rates $r$ are initialized with a time constant of 0.1\,s and are learned thereafter. A packet without events advances the state by its physical duration through transport and retention alone, since $S_k\equiv0$ in Eq.~\eqref{eq:spatial-update}.

\paragraph{Pyramid and readout (Sec.~\ref{sec:attention}, Eqs.~\eqref{eq:local-attn} and~\eqref{eq:depth-head}).}
Every pyramid level is computed from the current state $M_k$ by average pooling by the factor $s$ followed by the learned block of Table~\ref{tab:supp:settings}, and no scale carries its own recurrent memory. The decoder fuses the levels from coarse to fine with bilinear upsampling, so each finer descriptor combines evidence at its own resolution with context summarized at larger scales. The query $q_i'$ of Eq.~\eqref{eq:local-attn} is a 32-dimensional projection (LayerNorm, linear, GELU) of $q_i$, the normalized coordinates, $\log(1+\mathrm{age}_{\mathrm{ms}})$ of the physical age $b_k-t_i$ in milliseconds, $\tau_i$ and $p_i$. The keys $k_{ij}$ are the decoded descriptors of the four scales, sampled at the event and one cell to its left, right, top and bottom, each plus a learned scale embedding; keys outside the image are masked. $W_K$ and $W_V$ are $32\times32$, $W_Q$ is the identity and $d_a=32$. The context $c_i$ is fused with $q_i'$ by a residual linear layer and a residual MLP, and a linear layer gives $\ell_i$ of Eq.~\eqref{eq:depth-head}. %

\paragraph{Target-time supervision (Sec.~\ref{sec:supervision}, Eqs.~\eqref{eq:warp} and~\eqref{eq:silog}).}
For an interval $[t_0,t_r]$ that ends at the reference time, the renderer splats each transported depth bilinearly and averages, $\widehat D_r(u)=\sum_i w_i\beta_{u,i}\tilde d_i^r/\sum_i w_i\beta_{u,i}$, with the bilinear weight $\beta_{u,i}$ of $\tilde u_i^r$ at pixel $u$ and the recency weight $w_i=\exp\bigl(-(t_r-t_i)/(0.2\,(t_r-t_0))\bigr)$, so every contributing event receives a gradient. %
A transported event contributes only if its depth $\tilde d_i^r$ is positive and finite and its projection $\tilde u_i^r$ lies inside the reference image, and a pixel enters $\Omega_r$ only if its rendered and reference depths are both valid, with the reference depth between 0.25 and 30\,m. Training minimizes Eq.~\eqref{eq:silog} with unit weight; each window is unrolled through its six packets from a reset state before the update.

\paragraph{Encoder initialization (Sec.~\ref{sec:distillation}).}
In stage~1 (Table~\ref{tab:supp:settings}), the event encoder is trained for 20 epochs to match the teacher features of Sec.~\ref{sec:distillation}; the spatial-relation term is the mean squared difference between the token cosine-similarity matrices of student and teacher after pooling both to $15\times20$. %
In stage~2, the depth model loads the compatible encoder tensors of the distilled checkpoint, the local-attention head of Sec.~\ref{sec:attention} is newly initialized, and a new optimizer is created. The learning rates follow a plateau schedule on validation AbsRel, which is also the checkpoint-selection criterion of Sec.~\ref{sec:exp-protocol}.

\section{Experimental details for Sec.~\ref{sec:experiments}}
\label{sec:supp-experiments}

\subsection{Input intervals and event subsampling (Sec.~\ref{sec:exp-protocol})}
\label{sec:supp-sampling}
\paragraph{Intervals.}
Candidate intervals span adjacent depth annotations of the MVSEC outdoor recordings at the native $260\times346$ resolution; an interval that fails the criteria of Sec.~\ref{sec:exp-protocol} is removed rather than moved to a nearby annotation. The endpoint poses and the pose at every event timestamp used in Eq.~\eqref{eq:warp} are interpolated at the exact timestamps. The same interval identifiers and endpoint times are used for training, rendering, scoring and baseline pairing.

\paragraph{Subsampling to 7,500 events.}
Every accepted interval holds at least 10,000 events and is reduced to the 7,500 of Sec.~\ref{sec:exp-protocol} before it is divided into packets, balancing it in space and time jointly. Its events are grouped into cells formed by 30 equal time bins between the first and last event and a $16\times12$ grid of image cells. The sampler finds the largest quota $\kappa$ for which taking $\min(n_c,\kappa)$ events from every occupied cell $c$, holding $n_c$ events, fits the budget, gives each remaining slot to a different randomly chosen cell that still holds unselected events, and draws the events of every cell uniformly without replacement. The retained events keep their exact coordinates, timestamps and polarities and are returned in time order. Only then is the interval divided into the $K=6$ equal-duration packets of Sec.~\ref{sec:method-formulation}, so each packet receives the retained events of its own time span.

\begin{figure*}[t]
    \centering
    \includegraphics[width=\textwidth]{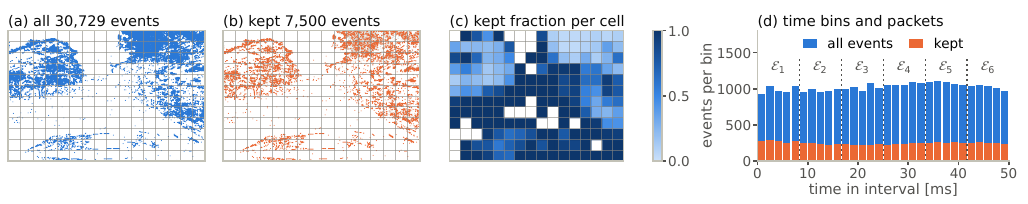}
    \caption{Fixed-count input of one 50\,ms \texttt{outdoor\_day1} interval. (a) All 30{,}729 events and the $16\times12$ sampling grid. (b) The 7{,}500 kept events. (c) Kept fraction per grid cell, from 4.6\% in dense cells to all events in sparse ones (white: no events). (d) Events per sampling time bin before and after selection, and the six packets $\mathcal{E}_k$. Each of the 3{,}167 occupied time--grid cells keeps $\min(n,2)$ of its $n$ events; 1{,}815 cells keep one more.}
    \label{fig:supp:subsampling}
\end{figure*}

Figure~\ref{fig:supp:subsampling} applies the sampler to a Day~1 interval with 30,729 events, the accepted interval closest to the 75th percentile of event counts. Its events occupy 3,167 of the 5,760 time--space cells. The quota is two events per cell, and the remaining 1,815 slots add a third event to as many cells; 1,245 cells are kept whole. Textured regions are thinned most: the 19 busiest of the 192 image cells hold 57.7\% of all events but 21.8\% of the retained ones (Fig.~\ref{fig:supp:subsampling}c), so sparse structure elsewhere in the image remains represented. In time, each bin keeps close to its share of the interval, and the six packets receive 1,155--1,386 events (Fig.~\ref{fig:supp:subsampling}d).

\subsection{Scoring and metrics (Sec.~\ref{sec:exp-protocol}, Tables~\ref{tab:ch3:depth-per-seq-mvsec} and~\ref{tab:ch3:depth-range-mae})}
\label{sec:supp-metrics}
\paragraph{Scored pixels.}
Sec.~\ref{sec:exp-protocol} gives the upper limits of the evaluated depth; the lower limit is 2\,m. EventSSM's per-event depths are transported to the reference camera with the interpolated annotated poses of Eq.~\eqref{eq:warp} and rendered as in training. For each reference frame, the scored set $\Omega$ contains the pixels whose reference depth lies between 2 and 30\,m, that EventSSM's rendered prediction covers, and at which every compared method gives a finite positive depth, including the restriction imposed by EReFormer's fixed $224\times224$ window. Every method is scored on this set, and errors are pooled over the scored pixels of each sequence, so MAE@30 in Table~\ref{tab:ch3:depth-range-mae} equals the MAE of Table~\ref{tab:ch3:depth-per-seq-mvsec}.

\paragraph{Metrics.}
With reference depth $d_u$ and prediction $\hat d_u$ at the scored pixels $u\in\Omega$, AbsRel and SqRel average $|\hat d_u-d_u|/d_u$ and $(\hat d_u-d_u)^2/d_u$ over $\Omega$, MAE and RMSE are the mean absolute and root-mean-square residuals in metres, $\delta_j$ is the percentage of pixels with $\max(\hat d_u/d_u,d_u/\hat d_u)<1.25^j$, and MAE@$r$ evaluates MAE on the subset of $\Omega$ with $d_u\le r$.

\subsection{Baselines (Sec.~\ref{sec:exp-depth}, Table~\ref{tab:ch3:depth-per-seq-mvsec})}
\label{sec:supp-baselines}
E2Depth~\cite{Hidalgo2020E2Depth} uses the released model trained on DENSE and MVSEC, DepthAnyEvent-R~\cite{Bartolomei2025DepthAnyEvent} the released recurrent model fine-tuned on MVSEC, and ScaleEvent~\cite{Ch5Chen2026ScaleEvent} the released MVSEC depth model with a ViT-S encoder, run at a $640\times480$ input. No trained EReFormer~\cite{Liu2024RecurrentTransformers} model is released, so it is trained on \texttt{outdoor\_day2} with its unmodified official code. Every baseline receives all natural events of the same interval in its own input representation and predicts one depth map at the reference time, and recurrent baselines start each interval from a reset state, as EventSSM does.

\subsection{Ablation protocols and results (Secs.~\ref{sec:exp-ablation}--\ref{sec:exp-components})}
\label{sec:supp-ablations}
\paragraph{Temporal cores (Sec.~\ref{sec:exp-ablation}, Table~\ref{tab:ch3:spatial-core-ablation}).}
The spatial RNN, spatial LSTM and bi-Mamba rows replace the update of Eq.~\eqref{eq:spatial-update} by a recurrent cell applied at every pixel of the state, with transitions shared across pixels ($1\times1$ convolutions), while keeping the learned transport of Eq.~\eqref{eq:transport}, the 64-channel encoder width and the event-conditioned readout of Sec.~\ref{sec:attention}. %
Every row, including EventSSM, is trained for 15 epochs without the initialization of Sec.~\ref{sec:distillation}, and its checkpoint is selected by validation AbsRel. Packet times are measured on one RTX 2080 Ti in FP32 with batch size one, in a benchmark separate from that of Table~\ref{tab:ch3:depth-baseline-runtime-same-node}.

\paragraph{Packet count (Sec.~\ref{sec:exp-packets}, Fig.~\ref{fig:ch3:packet-count-day1}).}
The fixed-weight curves evaluate the $K=6$ model on the same windows divided into $K\in\{1,3,6,12,24\}$ equal-duration packets; the per-$K$ curves train the full model separately for each $K$ with a 20-epoch budget and select it by Day~1 AbsRel. All points are scored on the same 3,066 Day~1 windows, on the pixels valid for every configuration, 16.98\% of the eligible reference pixels between 2 and 30\,m, with the state reset before each window. Across $K$, the fixed-weight MAE stays between 3.650 and 3.669\,m and its AbsRel between 0.311 and 0.315; the per-$K$ models reach 3.607--3.617\,m and 0.330--0.334.

\paragraph{Component ablations (Sec.~\ref{sec:exp-components}, Table~\ref{tab:ch3:component-ablation}).}
The errors of Table~\ref{tab:ch3:component-ablation} are averaged over 3,066 Day~1 windows and over 7,667 windows pooled from the three night sequences, with reference depths between 2 and 30\,m on each model's own rendered support, and every variant is trained with two seeds.

\begin{table}[tb]
    \centering
    \caption{Component ablations of Table~\ref{tab:ch3:component-ablation}: absolute relative error and mean absolute error after per-frame median scaling, averaged over evaluation windows. Values are the mean $\pm$ half-range over the same two training seeds. Each group starts with EventSSM (full); lower is better, best per column within each group in \textbf{bold}.}
    \label{tab:supp:component-ablation-relative}
    \begin{adjustbox}{max width=\linewidth}
        \begin{tabular}{@{}lcccc@{}}
            \toprule
            & \multicolumn{2}{c}{AbsRel $\downarrow$} & \multicolumn{2}{c}{Median-scaled MAE (m) $\downarrow$} \\
            \cmidrule(lr){2-3}\cmidrule(l){4-5}
            Variant & Day~1 & Nights & Day~1 & Nights \\
            \midrule
            \multicolumn{5}{@{}l}{\emph{Spatial state (Sec.~\ref{sec:spatial-state})}} \\
            EventSSM (full) & 0.295 $\pm$ 0.003 & 0.321 $\pm$ 0.002 & 3.302 $\pm$ 0.022 & 3.372 $\pm$ 0.006 \\
            No transport & 0.295 $\pm$ 0.001 & 0.322 $\pm$ 0.007 & 3.315 $\pm$ 0.007 & 3.372 $\pm$ 0.013 \\
            Per-update retention & 0.293 $\pm$ 0.003 & 0.325 $\pm$ 0.001 & 3.362 $\pm$ 0.042 & 3.465 $\pm$ 0.056 \\
            No forgetting & 0.293 $\pm$ 0.001 & 0.320 $\pm$ 0.002 & 3.302 $\pm$ 0.010 & 3.367 $\pm$ 0.007 \\
            No input gate & \textbf{0.291} $\pm$ 0.002 & \textbf{0.311} $\pm$ 0.004 & \textbf{3.282} $\pm$ 0.006 & \textbf{3.316} $\pm$ 0.009 \\
            No support mask & 0.295 $\pm$ 0.004 & 0.320 $\pm$ 0.001 & 3.323 $\pm$ 0.017 & 3.387 $\pm$ 0.025 \\
            \midrule
            \multicolumn{5}{@{}l}{\emph{Depth head (Sec.~\ref{sec:attention})}} \\
            EventSSM (full) & 0.295 $\pm$ 0.003 & 0.321 $\pm$ 0.002 & \textbf{3.302} $\pm$ 0.022 & 3.372 $\pm$ 0.006 \\
            Average-pooled scales & 0.296 $\pm$ 0.000 & \textbf{0.320} $\pm$ 0.002 & 3.304 $\pm$ 0.001 & \textbf{3.352} $\pm$ 0.007 \\
            Centre-sample readout & \textbf{0.292} $\pm$ 0.002 & 0.323 $\pm$ 0.001 & 3.319 $\pm$ 0.051 & 3.433 $\pm$ 0.063 \\
            \midrule
            \multicolumn{5}{@{}l}{\emph{Supervision (Sec.~\ref{sec:supervision})}} \\
            EventSSM (full) & 0.295 $\pm$ 0.003 & \textbf{0.321} $\pm$ 0.002 & \textbf{3.302} $\pm$ 0.022 & \textbf{3.372} $\pm$ 0.006 \\
            No target-time warp & \textbf{0.293} $\pm$ 0.000 & 0.322 $\pm$ 0.001 & 3.311 $\pm$ 0.004 & 3.378 $\pm$ 0.007 \\
            \bottomrule
        \end{tabular}
    \end{adjustbox}
\end{table}

Table~\ref{tab:supp:component-ablation-relative} adds AbsRel and the MAE for the same runs and windows, to separate effects on metric scale from effects on relative depth structure. The MAE rescales each frame by the ratio of the reference and predicted median depths over its scored pixels, so it measures the error left once the frame's global scale is corrected. Per-update retention keeps its night increase after median scaling (0.093\,m, 2.8\%), with both seeds above both full-model seeds, and raises the Day~1 scaled error by 0.060\,m (1.8\%). Without the input gate, the night AbsRel falls from 0.321 to 0.311 and the night scaled error by 0.056\,m (1.7\%), with both seeds below those of the full model, whereas its unscaled night error is 0.004\,m higher. Every other variant changes AbsRel by at most 0.004. Average-pooled scales change the scaled error by $+0.002$\,m on Day~1 and $-0.020$\,m at night, and training without the target-time warp raises it by 0.009 and 0.006\,m.

\end{document}